\documentclass[runningheads]{llncs}
\usepackage{graphicx}

\usepackage[table]{xcolor}
\usepackage{latexsym}
\usepackage{multirow}
\usepackage{adjustbox}
\usepackage{pifont}
\usepackage{cleveref}
\usepackage{makecell}
\usepackage[lofdepth,lotdepth]{subfig}
\usepackage[marginal]{footmisc}

\usepackage{marvosym}
\begin{document}
\title{Rethinking the Value of Gazetteer in Chinese Named Entity Recognition}

\author{Qianglong Chen\textsuperscript{1}, Xiangji Zeng\textsuperscript{1}, Jiangang Zhu\textsuperscript{2}, Yin Zhang\textsuperscript{1(\Letter)\thanks{Corresponding author: Yin Zhang.}}, Bojia Lin\textsuperscript{2}, Yang Yang\textsuperscript{2}, Daxin Jiang\textsuperscript{2}
}
\institute{\textsuperscript{1}College of Computer Science and Technology, Zhejiang University, China \\\email{\{chenqianglong,zengxiangji,zhangyin98\}@zju.edu.cn}
\\\textsuperscript{2}Microsoft STCA \\\email{\{jiangazh,bojial,yayan,djiang\}@microsoft.com}}

% \author{}
% \institute{}

\authorrunning{Q. Chen, X. Zeng, et al.}

%
\maketitle
\footnote{Q. Chen and X. Zeng —— Equal contribution.}
\begin{abstract}
Gazetteer is widely used in Chinese named entity recognition (NER) to enhance span boundary detection and type classification. 
However, to further understand the generalizability and effectiveness of gazetteers,
the NLP community still lacks a systematic analysis of the gazetteer-enhanced NER model.
In this paper, we first re-examine the effectiveness several common practices of the gazetteer-enhanced NER models and carry out a series of detailed analysis to evaluate the relationship between the model performance and the gazetteer characteristics, which can guide us to build a more suitable gazetteer. The findings of this paper are as follows: (1) the gazetteer improves most of the situations that the traditional NER model datasets are difficult to learn. (2) the performance of model greatly benefits from the high-quality pre-trained lexeme embeddings. (3) a good gazetteer should cover more entities that can be matched in both the training set and testing set.
\keywords{Gazetteer, Chinese Named Entity Recognition, Knowledge Enhancement}
\end{abstract}
\section{Introduction}
Gazetteer, also known as the entity dictionary, has been mentioned in various literature about its importance to Chinese NER, such as solving error propagation in Chinese NER~\cite{zhang_chinese_2018,sui_leverage_2019,gui-etal-2019-lexeme,ma_simplify_2020} and integrating rich external knowledge into NER~\cite{ding_neural_2019,lin_gazetteer-enhanced_2019,liu_towards_2019}. 
%GENER的流程
Previous studies have proved that compared with the traditional NER model, the gazetteer-enhanced NER model can obtain external boundary and type information from gazetteer knowledge to boost the performance. 
The typical process of a gazetteer-enhanced NER consists of four stages: (1) Collecting word entries from knowledge bases to construct a gazetteer. The word entries could be named entities or arbitrary phrases. (2) Conducting entry matching on the target sentence. (3) Encoding sentence and gazetteer knowledge with various encoders. (4) Extracting entities through decoding context-aware representation.

Although several gazetteer-enhanced NER models have been proposed in recent years, there are still several key questions that need to be answered. 
Firstly, several measures~\cite{fu_rethinking_2020} 
have be developed to analyze the generalization behavior of existing NER models.
However, the community is still lacking a detailed and unified evaluation for inspecting the relationship between the gazetteer-enhanced NER model's performance and its three indispensable components (gazetteer, dataset, and model) while now heavily depending on a holistic metric (F1 score). 
Secondly, the community is growing too fast to lack a comprehensive and systematic empirical study on gazetteer-enhanced NER models for reviewing the promising works in the past years and rethinking the pros and the cons.

\label{questions}
To solve the above problems, in this work, we design several experiments which are committed to answering three questions as follows:
\begin{itemize}
    \item \textbf{Q1:} Does gazetteer only have a positive impact on the NER model, or it also has a negative impact?
    \item \textbf{Q2}: Can the gazetteer improve the performance of the NER models based on the pre-trained language model ? 
    \item \textbf{Q3}: What type of gazetteer is the best gazetteer for improving performance?
\end{itemize}
%gazetteer对NER是否有用，这个问题太粗糙, gazetteer肯定是有用的
%gazetteer能不能在PLM为back-bone的模型上获得提升，其实Q1&Q2是一样的，没什么意义
%这个问题有些意义，什么类型的词汇更有效
%复现了已有的工作
To answer \textbf{Q1}, we reproduce three promising models and conduct several experiments on five datasets with three gazetteers. 
While almost all research efforts ignore whether the gazetteer is still helpful to the models enhanced by large scale pre-training language models, to figure out \textbf{Q2}, we modify the original static embedding as the pre-trained language model and conduct several new experiments on these new models. 
Additionally, current research works do not investigate the relationship between the model performance and the gazetteer characteristics, such as gazetteer size, pre-trained lexeme embedding, and lexeme type.
To further explore \textbf{Q3}, we design experiments to investigate these relationship, which can provide interpretable results for better understanding the causality behind the entity extraction and guide us to build a more suitable gazetteer. 
Our code is available at this URL\footnote{\url{https://github.com/knowledgeresearch/kaner}}.

The main contributions of this paper are summarized as follows: 
\begin{itemize}
    \item  We first re-examine some promising models with three gazetteers on five datasets with analysis through a fair and comparable environment.
    \item To further evaluate the effectiveness of gazetteer in pre-trained language models, we modify the gazetteer-enhanced model to adapt the existing pre-trained language models and make detailed analysis.
    \item Moreover, we conduct a series of explorations to uncover the relationship between the model performance and the gazetteer characteristics, which would bring some new insights for future work.
\end{itemize}

% \section{Preliminary}
\section{Task Definition}
In general, the gazetteer-enhanced named entity recognition (GENER) task consists of three indispensable components: a gazetteer \(\mathcal{G} = (l_1, l_2, ..., l_M)\), a dataset \(\mathcal{D} = \{(X, Y)\}\), and a GENER model \(\mathcal{M}\). 
Here, we denote \(X = \{x_1, x_2, ..., x_N\}\) as an input sequence and \(Y = \{y_1, y_2, ..., y_N\}\) as the corresponding output tags. 
To obtain external lexeme information \(L\), we adopt lexeme matching on the input sequence \( X \) with the gazetteer \(\mathcal{G}\). 
The goal of the task is to estimate the conditional probability \(P(Y|X, L)\) with model \(\mathcal{M}\).

\begin{table}[t]
\begin{center}
  \begin{adjustbox}{max width=\linewidth}
  \begin{tabular}{l | c c | c c c}
    \Xhline{3\arrayrulewidth}
        \multirow{2}{*}{\textbf{Model}} & \multicolumn{2}{c}{\textbf{Token Embedding}} & \multicolumn{3}{|c}{\textbf{Gazetteer}} \\
        \cline{2-6}
        & {\small Gigaword} & {\small BERT} & {\small Gigaword} & {\small SGNS} & {\small TEC} \\
    \Xhline{3\arrayrulewidth}
    \textit{BLCRF} & \ding{51} & \ding{53}  & \ding{53}  & \ding{53} & \ding{53} \\
    \textit{PLMTG} & \ding{53} & \ding{51} & \ding{53} & \ding{53} &  \ding{53}\\
    \textit{SES} & \ding{51} &\ding{53} & \ding{51} & \ding{51} & \ding{51} \\
    \textit{SES\(^{\dag}\)} &\ding{53} & \ding{51} & \ding{51} & \ding{51} & \ding{51} \\
    \textit{CGN} & \ding{51} & \ding{53}& \ding{51} & \ding{51} & \ding{51} \\
    \textit{CGN\(^{\dag}\)} & \ding{53}& \ding{51} & \ding{51} & \ding{51} & \ding{51} \\
    \textit{MDGG} & \ding{51} & \ding{53}& \ding{51} & \ding{51} & \ding{51} \\
    \textit{MDGG\(^{\dag}\)} & \ding{53}& \ding{51} & \ding{51} & \ding{51} & \ding{51} \\
    \Xhline{3\arrayrulewidth}
    \end{tabular}
    \end{adjustbox}
\end{center}
\caption{Models with Different Architectures. All our models are based on character-level token embeddings.\label{tab:method_overview}}
\end{table}

\section{Model}
\label{model}

% It will be challenging to reproduce all GENER models well and manage amounts of experiments under limited resources. Hence, we put all our energy into a limited number of models. It would be interesting to select which model to reproduce. 
\subsection{GENER Model Selection \& Reproduction}
For the GENER model selection, we follow some rules as below.
Firstly, the results of the model reported by the original paper should be \textit{promising} enough. 
Secondly, the model's architecture should be \textit{representative} enough. 
Considering the two points, we divide the existing models into two categories: token embedding modification and context encoder modification.
The former integrates extra lexeme information with the token embedding. The latter modifies the context encoder to adapt extra lexeme information. 
Therefore, we select a model (SES) from the former category and two models (CGN, MDGG) from the latter category to investigate this area's progress. 
%Due to the time limitation, experiments on more models are left in future work.
As we can see in Table ~\ref{tab:method_overview}, we reproduce three promising GENER models, namely SES, CGN, and MDGG. The SES model~\cite{ma_simplify_2020} integrates external lexeme information into token embeddings. The CGN model~\cite{sui_leverage_2019} and the MDGG model~\cite{ding_neural_2019} treat the matched lexemes and the original token sequence as a graph, then encode the graph by graph neural networks. 

\subsection{Model Replacement with Pre-trained Language Model}
To assess some components' role, we vary these models in terms of two aspects: different token embeddings or gazetteers. To explore the question \textbf{Q2}, we use BERT~\cite{devlin-etal-2019-bert} to replace the original static token embedding. 
% In this part, we take BERT~\cite{devlin-etal-2019-bert} as a backbone to analyze the pre-trained language model's possible improvement. 
%To keep the integrity and learned prior knowledge of BERT, we directly replace the GENER model's token embedding module with BERT. 
Specifically, we let BERT take the original input and regard BERT's final output as the token embeddings of the GENER model. When training the GENER model, we set the BERT part to have a low learning rate while setting a high learning rate for the remaining part of the GENER model. %In our experiments, we follow the previous experimental procedures and environment, except for the number of training epochs, because these improved GENER models need more iterations to converge.
Additionally, we also implement two baseline NER models: BLCRF and PLMTG. The BLCRF model~\cite{panchendrarajan-amaresan-2018-bidirectional} consists of a bidirectional LSTM and a CRF layer. The PLMTG model consists of a BERT and a MLP layer. 
All our models are based on character-level token embeddings.

\section{Experiments}
% In this section, we dedicate to design experiments for answering the three questions. It's worth noting that our experimental results may \textit{differ} from the results reported by the original papers because we need to ensure the consistency of the environment for \textit{fair comparison}.
\subsection{Gazetteer \& Dataset}
\subsubsection{Gazetteer}
\begin{table}[t]
\begin{center}
  \begin{adjustbox}{max width=\linewidth}
  \begin{tabular}{l | l l c}
    \Xhline{3\arrayrulewidth}
        \textbf{Gazetteer} & \textbf{Num.} & \textbf{Dim.} & \textbf{Pre-trained} \\
    \Xhline{3\arrayrulewidth}
        TEC & 61400 & 50 & \ding{53} \\
        Gigaword & 704368 & 50 & \ding{51} \\
        SGNS & 1292607 & 300 & \ding{51} \\
    \Xhline{3\arrayrulewidth}
    \end{tabular}
    \end{adjustbox}
\end{center}
\caption{
Statistics of Gazetteers. \textit{Num.} represents the number of lexeme in a gazetteer. \textit{Dim.} represents the dimension of lexeme embedding. \textit{pre-trained} represents whether the lexeme embeddings are pre-trained, such as Word2vec~\cite{DBLP:journals/corr/abs-1301-3781}, GloVe~\cite{pennington-etal-2014-glove}, etc. 
%If they are not pre-trained, all lexeme embeddings will be randomly initialized. 
\label{tab:statistic_of_gazetteers}
}
\end{table}
We choose three different gazetteers: Gigaword, SGNS, and TEC, to verify the effectiveness of gazetteer in the NER task.
The Gigaword gazetteer~\cite{yang-etal-2017-neural-word} contains lots of words from the word segmentator, pre-trained embeddings and character embeddings, which is trained from the Chinese Gigaword corpus\footnote{https://catalog.ldc.upenn.edu/LDC2011T13}.
The SGNS gazetteer~\cite{P18-2023} is
trained from Wikipedia and News and contains lots of words from the word segmentator and pre-trained embeddings.
The TEC gazetteer~\cite{ding_neural_2019}, collected from the e-commerce domain, contains product and brand names only. The statistics of gazetteers are shown in Table~\ref{tab:statistic_of_gazetteers}.

\subsubsection{Dataset}
\begin{table}[t]
\begin{center}
  \begin{adjustbox}{max width=\linewidth}
  \begin{tabular}{l | l l l l}
    \Xhline{3\arrayrulewidth}
        \textbf{Dataset} & \textbf{Total} & \textbf{Train} & \textbf{Dev} & \textbf{Test} \\
    \Xhline{3\arrayrulewidth}
        WeiboNER & 1890 & 1350 & 270 & 270 \\
    \hline
        ResumeNER & 4761 & 3821 & 463 & 477 \\
    \hline
        ECommerce & 4987 & 3989 & 500 & 498 \\
    \hline
        OntoNotes & 49306 & 39446 & 4930 & 4930 \\
    \hline
        MSRANER & 50729 & 41728 & 4636 & 4365 \\
    \Xhline{3\arrayrulewidth}
    \end{tabular}
    \end{adjustbox}
\end{center}
\caption{Statistics of Selected Datasets.  %Each number represents the number of data samples in a sub-dataset.
\label{tab:statistic_of_datasets}}
\end{table}

To make the results more convincing, we select five Chinese NER datasets: WeiboNER, ResumeNER, MSRANER, OntoNotes, and ECommerce, which are distributed from a wide range of domains, including social media, financial articles, news, and e-commerce. The WeiboNER dataset~\cite{peng-dredze-2015-named}, annotated from Weibo\footnote{https://weibo.com/} messages, contains both named and nominal mentions. The ResumeNER dataset~\cite{zhang_chinese_2018}, collected from Sina Finance\footnote{https://finance.sina.com.cn/stock/}, consists of senior executives' resumes from listed companies in the Chinese stock market. The MSRANER dataset~\cite{gao-etal-2005-chinese} is collected from the news domain. The OntoNotes dataset\footnote{https://catalog.ldc.upenn.edu/LDC2013T19/}, based on the OntoNotes project, is annotated from a large corpus comprising various text genres, such as news, conversational telephone speech, etc. The ECommerce dataset~\cite{ding_neural_2019}, collected from the e-commerce domain, contains lots of product descriptions. The statistics of datasets are shown in Table~\ref{tab:statistic_of_datasets}.

\definecolor{LightBlue}{RGB}{231, 245, 255}
\begin{table}[t]
    \centering  
    \begin{adjustbox}{max width=\linewidth}
    \begin{tabular}{l l | l l l l l}
    \Xhline{3\arrayrulewidth}
        \rowcolor{LightBlue} \textbf{Model} & \textbf{Gazeteer} & \textbf{WeiboNER} & \textbf{ResumeNER} & \textbf{ECommerce} & \textbf{MSRANER} & \textbf{OntoNotes} \\
    \Xhline{3\arrayrulewidth}
        \multicolumn{7}{c}{\textit{Vanilla Model}} \\
    \Xhline{2.5\arrayrulewidth}
        BLCRF & N/A & 50.37 & 92.91 & 64.99 & 84.48 & 90.07  \\
    \Xhline{2.5\arrayrulewidth}
        \multirow{3}{*}{SES} & Gigaword & 53.04 & 93.37 & 69.48 & 88.05 & 90.23  \\
        & SGNS & \textbf{57.22} \color{blue}{(+6.85)} & 93.45 & \textbf{71.90} \color{blue}(+0.691) & \textbf{90.87}(+6.39) & \textbf{90.49}\color{blue}(+0.42) \\
        & TEC & 52.37 & 92.94 & 65.19 & 85.47 & 89.96  \\
    \hline
        \multirow{3}{*}{CGN} & Gigaword & 51.12 & 92.72 & 63.91 & 86.73 & 88.23  \\
        & SGNS & 54.24 & 92.22 & 65.89 & 89.10 & 88.12  \\
        & TEC & 49.58 & 92.82 & 64.37 & 85.16 & 88.15 \\
    \hline
        \multirow{3}{*}{MDGG} & Gigaword & 53.16 & 93.86 & 70.07 & 76.16 & 63.96 \\
        & SGNS & 56.20 & 93.83 & 71.38 & 76.09 & 63.48 \\
        & TEC & 52.88 & \textbf{93.87}\color{blue}(+0.96) & 69.48 & 84.37 & 72.26 \\
    \Xhline{2.5\arrayrulewidth}
        \multicolumn{7}{c}{\textit{Revised Model}} \\
    \Xhline{2.5\arrayrulewidth}
        PLMTG & N/A & 64.35 & 90.90 & 76.61 & 89.58 & 87.32  \\
    \Xhline{2.5\arrayrulewidth}
        \multirow{3}{*}{SES\(^{\dag}\)} & Gigaword & 67.00 & 94.91 & 81.05 & \textbf{94.18} \color{blue}(+4.6) & \textbf{70.66} \color{blue}(-16.66)\\
        & SGNS & 67.89 & 95.76 &  \textbf{81.32}\color{blue}(+4.71) & 93.10 & 57.56 \\
        & TEC & 41.38 & 95.74 & 79.36 & 93.30 & 45.90 \\
    \hline
        \multirow{3}{*}{CGN\(^{\dag}\)} & Gigaword & 45.50 & 91.53 & 55.23 & 81.36 & 69.65 \\
        & SGNS & 54.30 & 92.22 & 54.94 & 87.36 & 54.24  \\
        & TEC & 27.48 & 59.67 & 47.38 & 37.37 & 18.89 \\
    \hline
        \multirow{3}{*}{MDGG\(^{\dag}\)} & Gigaword & \textbf{68.41} \color{blue}(+4.06) & 96.11 & 80.32 & 90.95 & 65.46 \\
        & SGNS & 67.94 & 95.81 & 80.73 & 92.55 & 44.83 \\
        & TEC & 67.78 & \textbf{96.27} \color{blue}(+5.37)& 80.11 & 89.60 & 66.47\\
    \Xhline{3\arrayrulewidth}
    \end{tabular}
    \end{adjustbox}
    \caption{F1 Results on WeiboNER, ResumeNER, ECommerce, MSRANER, OntoNotes.}
    \label{tab:results_on_weiboner}
\end{table}

\begin{table}[t]
    \centering
    \begin{adjustbox}{max width=\linewidth}
    \begin{tabular}{l l | l l}
    \Xhline{3\arrayrulewidth}
        \textbf{Model} & \textbf{Gazeteer} & \textbf{\# Params.} & \textbf{\# Time} \\
    \Xhline{3\arrayrulewidth}
        BLCRF & N/A & 1.0 & 1.0 \\ 
    \hline
        \multirow{3}{*}{SES} & Gigaword & \(\times\) 27.1 & \(\times\) 1.3 \\
        & SGNS & \(\times\) 287.3 & \(\times\) 1.36 \\
        & TEC & \(\times\) 3.4 & \(\times\) 1.5 \\
    \hline
        \multirow{3}{*}{CGN} & Gigaword & \(\times\) 27.4 & \(\times\) 8.2 \\
        & SGNS & \(\times\) 286.8 & \(\times\) 8.5 \\
        & TEC & \(\times\) 3.8 & \(\times\) 5.3 \\
    \hline
        \multirow{3}{*}{MDGG} & Gigaword & \(\times\) 32.5 & \(\times\) 10.0 \\
        & SGNS & \(\times\) 291.9 & \(\times\) 11.1 \\
        & TEC & \(\times\) 11.0 & \(\times\) 6.7 \\
    \Xhline{3\arrayrulewidth}
    \end{tabular}
    \end{adjustbox}
    \caption{Model Complexity.}
    \label{tab:model_complexity}
\end{table}

% \subsubsection*{Exp-\rom{1}: Model Comparison}
\subsection{Fair Model Comparison Setting}
Since the hyperparameters in the GENER model are sensitive to different datasets, choosing suitable hyperparameters for each model for fair comparison is one of the most important problem.
% The biggest reproducibility problem we encounter is choosing suitable hyperparameters for each model under the condition that we have many experiments because hyperparameters in the GENER model are sensitive to different datasets. 
% To solve the above issue and ensure a fair comparison environment, 
For each model, we first adjust model hyperparameters on a small dataset (WeiboNER) and then reuse these hyperparameters on the other datasets and gazetteers. 
%The consistency of this operation is that a small dataset is usually more difficult than a big dataset due to spurious correlations~\cite{zeng-etal-2020-counterfactual}.
We use the micro-average F1 score to evaluate the model performance and
we also give the average training speed and the number of parameters to illustrate the model complexity.

%\paragraph{Analysis} 
\subsection{Experimental Results and Analysis}
\subsubsection{The Effectiveness of Gazetteer in GENER}
As shown in the part of the vanilla model of Table~\ref{tab:results_on_weiboner}, the GENER model is outstanding in the scenario where the baseline model BLCRF performs poorly. 
Specifically, the SES model with the SGNS gazetteer gets the best in most cases. The fly in the ointment is that it has a large amount of parameters (See Table~\ref{tab:model_complexity}), which mainly comes from the word embeddings of SGNS gazetteer.
However, compared to the CGN model and the MDGG model, it is still superior in model complexity. 
The computation time of these two models is mainly restricted by graph neural networks. 
Additionally, compared to the baseline model, the MDGG model performs relatively poorly on large datasets (MSRANER and OntoNotes) but performs better on small datasets (WeiboNER, ResumeNER, and ECommerce). The results demonstrate that the gazetteer is helpful in most cases. However, an improper gazetteer also brings a negative impact. 
% Our more comprehensive experiments suggest that \textit{when the baseline model performs poorly in some scenarios, only applying a suitable gazetteer will largely boost the performance}. 
%We will discuss how to choose a proper gazetteer in Section~\ref{answer_to_q3}.

\subsubsection{Improvement on Pre-trained Language Models}
\label{improve}
Although we observe the positive impact of GENER models in most scenarios, %there is a question that always haunts us in our minds. Since the pre-trained language models have gradually dominated the natural language processing community~\cite{DBLP:journals/corr/abs-2005-14165}, 
almost all existing GENER models ignore to explore whether these gazetteer-enhanced methods can boost the performance of pre-trained language models or not.
% \subsubsection*{Exp-\rom{2}: Contextual Embedding Enhancement}
Driven by \textbf{Q2}, we incorporate the pre-trained language model with selected GENER methods. 
% As we all know, large-scale pre-trained language models can effectively improve the performance of downstream tasks, such as T5~\cite{2020t5} and GPT-3~\cite{DBLP:journals/corr/abs-2005-14165}. 
% However, in most scenarios, we want to use a smaller model to solve simple tasks like the NER task because of the low cost of a smaller model. 
%\paragraph{Analysis} 
The results are shown in the part of the revised model of Table~\ref{tab:results_on_weiboner}. 
Surprisingly, compared to the baseline model PLMTG, the BERT-enhanced GENER model is still boosting performance a lot in most scenarios. 
Furthermore, experimental results demonstrate the effectiveness of gazetteer in improving model performance in most scenarios where the baseline model PLMTG has a relatively higher F1 score. 
Unfortunately, it seems difficult to learn well on OntoNotes (See Table~\ref{tab:results_on_weiboner}) and CGN model perform poorly in most cases. It may be due to inappropriate hyperparameter selection~\cite{li2019deepgcns}. 
Compared with the GENER model without BERT, the BERT-enhanced GENER model can further improve the performance in most cases, which is what we have expected.
Compared to the PLMTG model, the gazetteer enhanced model can achieve a better performance, which indicates that \textit{the smaller pre-trained language model is not the NER's ceiling}.

\begin{figure*}[t]
    \centering
    \includegraphics[width=\linewidth]{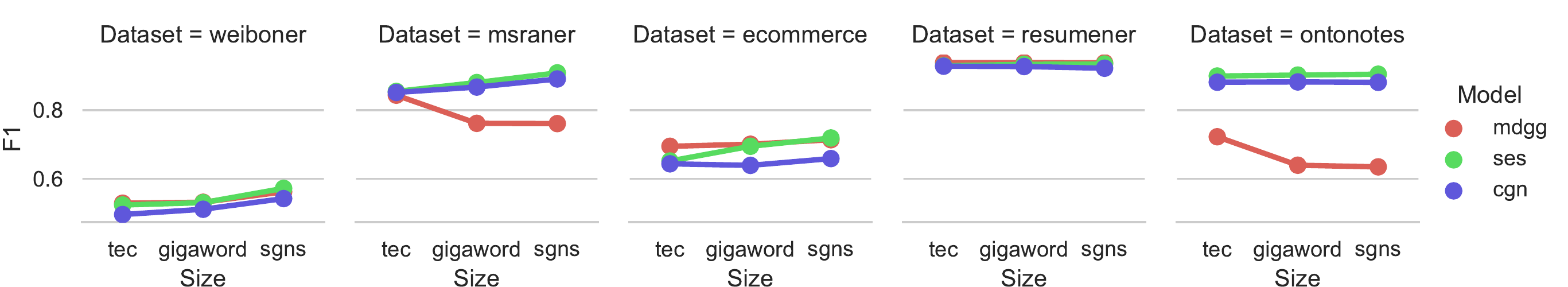}
    \caption{F1 score changes across different gazetteers. The three gazetteers are on the coordinate axis from left to right according to the gazetteer size (tec \(\rightarrow\) gigaword \(\rightarrow\) sgns).}
    \label{fig:gazsize_all}
\end{figure*}

\begin{figure}[t]
    \centering
    \includegraphics[width=\linewidth]{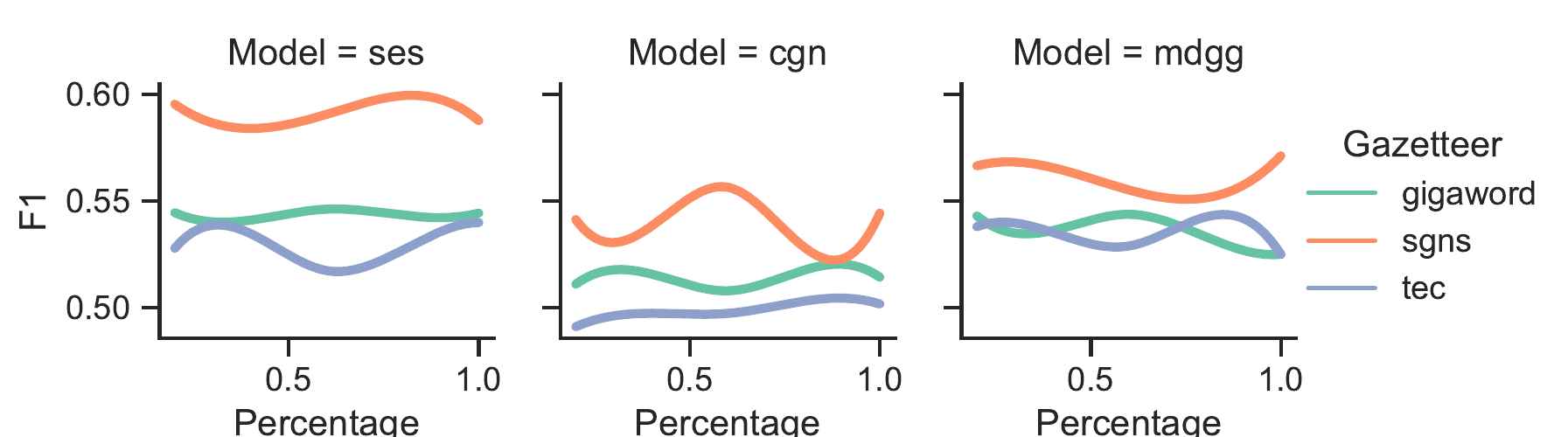}
    \caption{F1 score changes across the size of the gazetteer itself on WeiboNER.}
    \label{fig:gazsize_weiboner}
\end{figure}

\subsection{Proper Gazetteer Exploration}
\label{answer_to_q3}
While the GENER model has achieved impressive performance in most scenarios, what kind of gazetteer makes the model perform well still remains to be explored.
%This concern is related to \textbf{Q3}. It is useful to design a better gazetteer. 
To answer this question, we design a series of experiments from two aspects: the gazetteer itself, and the interaction between the gazetteer and dataset.

\subsubsection{Gazetteer Itself}
% In the next section, we will discuss more details on the lexeme.
We begin with characteristics of the gazetteer itself to figure out the relationship between the gazetteer and the model performance. 
For simplicity, we treat every lexeme in the gazetteer as an indivisible atom in this section. Therefore, the gazetteer's available characteristics are \textit{gazetteer size} and \textit{pre-trained lexeme embedding}. 
%\subsubsection*{Exp-\rom{3}: Gazetteer Size}
% \paragraph{Analysis} 
% Gazetteer size means how many lexemes a gazetteer contains. 
\paragraph{Gazetteer Size} 
As shown in Table~\ref{tab:statistic_of_gazetteers}, we list the number of lexeme for each gazetteer. 
% The SGNS gazetteer contains the most lexemes, while the TEC gazetteer contains the least lexemes. 
We first plot the relationship between gazetteer size and the model performance across the three gazetteers (See Figure~\ref{fig:gazsize_all}).
In most cases, the model performance would increase when the gazetteer size is getting bigger. 
However, there still exists a few cases that are opposite the former pattern. 
Then, we re-conduct experiments on WeiboNER with the three GENER models but only select a certain percentage of lexeme for each gazetteer. As shown in Figure~\ref{fig:gazsize_weiboner}, we observe that there is no obvious trend to prove that the larger the gazetteer is, the better the model performance will be. The results show that the model performance is not always positively correlated with the gazetteer size. 
\textit{A small gazetteer has the ability to play a big role}. It suggests that \textit{the gazetteer quality or suitability may influence the model performance.}

%\subsubsection*{Exp-\rom{5}: Pre-trained lexeme Embedding}
\paragraph{Pre-trained Lexeme Embedding}
\begin{table}[t]
    \centering
    \begin{adjustbox}{max width=\linewidth}
    \begin{tabular}{l l | l l l}
        \Xhline{3\arrayrulewidth}
            \textbf{Model} & \textbf{Gazetteer} & \textbf{F1} & \textbf{Precision} & \textbf{Recall} \\
        \Xhline{3\arrayrulewidth}
            \multirow{4}{*}{SES} & Gigaword & 53.04 & 59.65 & 47.78 \\
            & Gigaword\(^{\ddag}\) & 53.44 \color{red}{(+0.40)} & 60.79 & 47.68 \\
        \cline{2-5}
            & SGNS & 57.22 & 62.85 & 52.61 \\
            & SGNS\(^{\ddag}\) & 53.70 \color{red}{(-3.52)} & 60.72 & 48.16 \\
        \hline
            \multirow{4}{*}{CGN} & Gigaword & 51.12 & 59.46 & 44.93 \\
            & Gigaword\(^{\ddag}\) & 50.00 \color{red}{(-1.12)} & 61.42 & 42.17 \\
        \cline{2-5}
            & SGNS & 54.24 & 60.14 & 49.47 \\
            & SGNS\(^{\ddag}\) & 50.28 \color{red}{(-3.96)} & 58.44 & 44.15 \\
        \hline
            \multirow{4}{*}{MDGG} & Gigaword & 53.16 & 55.92 & 50.67 \\
            & Gigaword\(^{\ddag}\) & 53.25 \color{red}{(+0.09)} & 56.82 & 50.14 \\
        \cline{2-5}
            & SGNS & 56.20 & 57.29 & 55.17 \\
            & SGNS\(^{\ddag}\) & 54.20 \color{red}{(-2.00)} & 58.90 & 50.24 \\
        \Xhline{3\arrayrulewidth}
    \end{tabular}
    \end{adjustbox}
    \caption{An ablation study of pre-trained lexeme embedding on WeiboNER. \(\ddag\) denotes that the gazetteer has no pre-trained lexeme embeddings.}
    \label{tab:pre-trained_lexeme_embeddings}
\end{table}
% Pre-training is a dominant technique in the NLP community, which provides prior knowledge as a good start for the downstream task~\cite{tamborrino-etal-2020-pre}.
% Before, we have seen the powerful performance of the GENER model augmented by the pre-trained language model BERT in Section~\ref{improve}. 
% In this part, we aim to figure out that what role the pre-trained lexeme embedding plays. 
As shown in~\cref{tab:statistic_of_gazetteers,tab:results_on_weiboner}, the gazetteer with pre-trained lexeme embedding performs better than the gazetteer without one in most cases. 
In particular, we have observed that the SGNS gazetteer is almost better than the Gigaword gazetteer. 
The biggest difference between them is that the former has a better and bigger pre-trained lexeme embedding. 
We also conduct an experiment on WeiboNER, which is divided into two groups, with pre-trained lexeme embedding or without. As shown in Table~\ref{tab:pre-trained_lexeme_embeddings}, we observe that the experiments without pre-trained lexeme embedding, compared to the one with pre-trained lexeme embedding, almost show that the F1 score has dropped significantly. Even for the rest situations, they are only a trivial increase in model performance. The results demonstrate that \textit{a good pre-trained lexeme embedding will bring a positive impact on the model performance}.

\subsubsection{Gazetteer \(\times\) Dataset}
\begin{figure}[t]
    \centering
    \includegraphics[width=\linewidth]{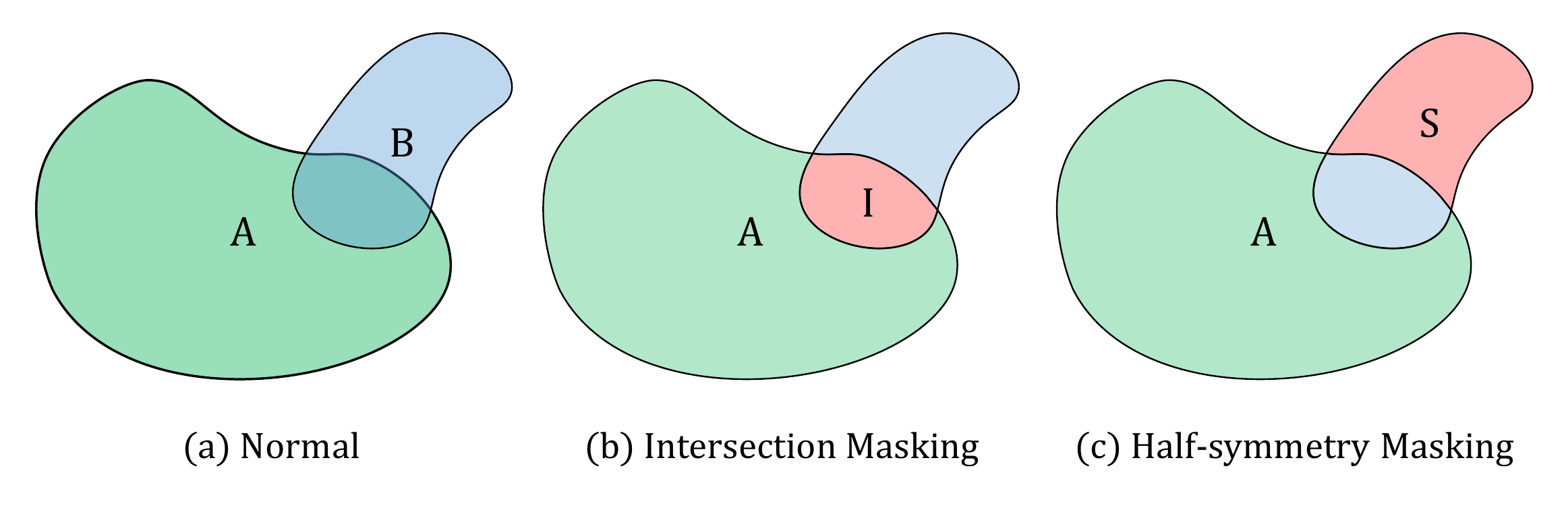}
    \caption{lexeme set masking in the test set. \(A\) denotes the matched lexeme set in the training set. \(B\) denotes the matched lexeme set in the test set.}
    \label{fig:test_masking}
\end{figure}

\begin{table}[t]
    \centering
    \begin{adjustbox}{max width=\linewidth}
    \begin{tabular}{l l | c c | c c}
        \Xhline{3\arrayrulewidth}
            \textbf{Dataset} & \textbf{Gazetteer} & \textit{I} & \textit{S} & \textit{E} & \textit{N} \\
        \Xhline{3\arrayrulewidth}
            \multirow{3}{*}{WeiboNER} & Gigaword & 3547 & 1698 & 220 & 5025 \\
            & SGNS & 3929 & 2003 & 278 & 5654 \\
            & TEC & 672 & 191 & 57 & 806 \\
        \hline
            \multirow{3}{*}{ResumeNER} & Gigaword & 2746 & 737 & 157 & 3326 \\
            & SGNS & 2715 & 868 & 276 & 3307 \\
            & TEC & 448 & 77 & 11 & 514 \\
        \hline
            \multirow{3}{*}{ECommerce} & Gigaword & 4609 & 1287 & 845 & 5051 \\
            & SGNS & 5266 & 1608 & 1082 & 5792 \\
            & TEC & 1513 & 473 & 953 & 1033 \\
        \hline
            \multirow{3}{*}{MSRANER} & Gigaword & 20887 & 4129 & 1877 & 23139 \\
            & SGNS & 21848 & 4370 & 2265 & 23953 \\
            & TEC & 1902 & 193 & 263 & 1832 \\
        \hline
            \multirow{3}{*}{OntoNotes} & Gigaword & 28439 & 1551 & 2293 & 27697 \\
            & SGNS & 27968 & 1561 & 2407 & 27122 \\
            & TEC & 2121 & 81 & 261 & 1941 \\
        \Xhline{3\arrayrulewidth}
    \end{tabular}
    \end{adjustbox}
    \caption{Statistics of Matched lexeme Set between the Dataset and the Gazetteer.}
    \label{tab:statistics_of_lexeme_set}
\end{table}

\begin{figure*}[t]
    \centering
    \includegraphics[width=\linewidth]{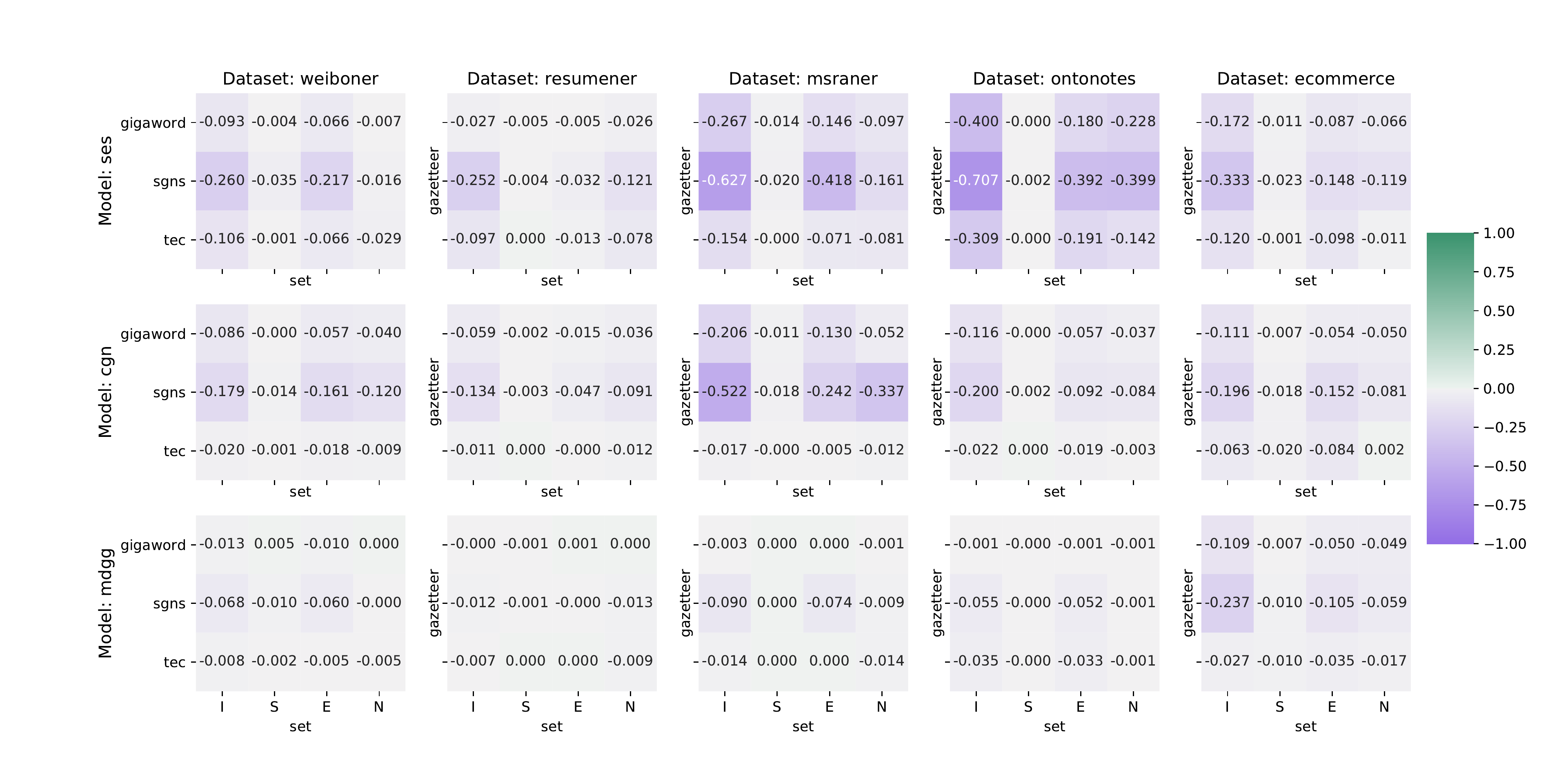}
    \caption{A heat map of causal effects on masking \(I\)\&\(S\) and \(E\)\&\(N\). It demonstrates the performance drop when masking different lexeme sets. The closer the color is to dark purple, the greater the degradation of model performance. This can be regarded as the lexeme set's importance.}
    \label{fig:masking_isen}
\end{figure*}

% If we understand the gazetteer only from the gazetteer itself, this way will be very rough, because the secret of the gazetteer is to interact with the dataset. Namely, only those matched lexemes in the dataset will play a role in the model. Here, the matched lexeme is gained by the exact match algorithm as the same as the original step in the three GENER models. Thus, 
In this section, we focus on the matched lexeme in a dataset. Specifically, these matched lexeme will be split into two different interpretable groups, the lexeme set \( A \) matched in the training set and the lexeme set \( B \) matched in the test set (See Figure~\ref{fig:test_masking}a). Our goal is to probe what kind of lexeme in the gazetteer would contribute more to the model performance.

%\subsubsection*{Exp-\rom{6}: lexeme Generalization}
\paragraph{Lexeme Generalization}
We firstly investigate the lexeme generalization ability: \textit{does the matched lexeme in the test set but unseen in the training set boost the model performance?}
Inspired by the concept of casual effects in causality~\cite{pearl2009causality}, we decide to mask a part of matched lexeme in the test for investigating how the model performance in a trained model will change. We further split \( B \) into two small sets. An intersection set between \( A \) and \( B \), denoted as
\begin{equation}
    I = A \cap B
\end{equation}
It represents the matched lexeme both shared in the training stage and the test stage (See Figure~\ref{fig:test_masking}b). A half-symmetry set in \( B \), denoted as
\begin{equation}
    S = (A \cup B) \setminus A
\end{equation}
It represents the matched lexeme in the test set but unseen in the training set (See Figure~\ref{fig:test_masking}c). 
We mask \( I \) and \( S \) separately to evaluate the model performance in the test set again. We calculate the causal effects using the difference between the F1 score masked in \( I \) and the original F1 score. This index indicates how much the model performance will decrease when masking \( I \). Similarly, we also have another experiment to evaluate the causal effects under the situation of masking \( S \).

%\subsubsection*{Exp-\rom{7}: Lexeme Type}
\paragraph{Lexeme Type}
% This experiment will tell us which type of lexeme is more effective in helping the GENER model recognize the boundaries of entities.
Secondly, we investigate that what type of lexeme the GENER likes. Here, the type means whether the lexeme is an entity annotated in the whole dataset. 
We split matched lexeme in the test set into two small sets, the lexeme set \( E \) which only includes entities, the lexeme set \( N \) which excludes all entities. 
We also mask the two lexeme sets separately to observe how the model performance will change in the test set.
Due to the space limitation, we plot all results of lexeme generalization and lexeme type together in the form of a heat map in Figure~\ref{fig:masking_isen}. 
Every column represents a dataset. Every row represents a model. The heat map shows the dropped model performance when masking a specific lexeme set. Table~\ref{tab:statistics_of_lexeme_set} demonstrates the count of four lexeme sets between datasets and gazetteers.

\begin{itemize}
    \item \textsc{Masking I\&S:} We can see the model performance largely decreases in most scenarios when masking \( I \), while almost unchanged or on reduces a little when masking \( S \). Although the count of \( S \) is less than the count of \( I \) in all combinations, the former still keeps a larger value. It suggests that it is difficult for the GENER model to generalize to the unmatched lexeme in the training stage.
    
    \item \textsc{Masking E\&N:} Contrary to the above experiment, the model performance decreases a lot both in masking \( E \) and \( N \), especially on the two big datasets, MSRANER and OntoNotes. However, except for a few limited cases, we can see that the most performance degradation occurs in masking \( E \). We also notice that the count of \( E \) is less than the count of \( N \) in all combinations. This phenomenon suggests that the entity helps the model to boost performance more.
\end{itemize}

After the above analysis, we summarized two points: (1) \textit{When building a gazetteer, we should consider more matched lexeme that cover both the training set and the test set}. (2) \textit{The more entities the gazetteer contains, the larger performance the model may boost}. 
% However, in real applications, we are unable to get the test set and tuning the lexeme based on the distribution of test sets. This challenge is currently ignored by the community. Therefore, it is important to develop a technique to solve this problem in the future.

% \subsubsection{Hard Cases}
% TODO
% Our work has connections to the following fields:

% \paragraph{Analysis} There are few studies towards a comprehensive analysis of the NER model performance~\cite{fu_rethinking_2020,fu-etal-2020-interpretable}. 

% \paragraph{Knowledge Integration}
\section{Related Works}
 Recent years, lots of methods have been developed for knowledge enhancement in named entity recognition. 
 These methods can be divided into two parts, supervised method and unsupervised method. 
 Most of works follow the supervised fashion. Some works attempt to modify the encoder to adapt the external gazetteer knowledge~\cite{zhang_chinese_2018,sui_leverage_2019,ding_neural_2019,gui_cnn-based_2019,zhou_chinese_2019,lin_gazetteer-enhanced_2019,liu_towards_2019,DBLP:journals/corr/abs-2004-06384,gui-etal-2019-lexeme}. 
Although these encoder-based models have achieved surprising results, they perform poorly in encoder transfer-ability and computational complexity. 
 Hence, another group of studies design methods for incorporating gazetteer information with token embedding~\cite{ma_simplify_2020,liu-etal-2019-encoding,li-etal-2020-flat,magnolini-etal-2019-use,rijhwani-etal-2020-soft,DBLP:journals/corr/abs-2004-04060,chen-etal-2020-improving}. 
 The unsupervised method mainly has two works, including PU learning~\cite{peng-etal-2019-distantly} and linked HMMs~\cite{Safranchik2020WeaklySS}. It is worth mentioning that \cite{williams-2019-neural} introduces lexeme knowledge to slot tagging.

\section{Conclusion}
In this paper, we first re-examine the effectiveness of several common practices gazetteer-enhanced NER model through a fair and comparable environment. 
To further evaluate the effectiveness of gazetteer in pre-trained language models, we modify the gazetteer enhanced model to adapt existing pre-trained language models. 
Moreover, we carry out a series of analysis of the relationship between the performance of model and the gazetteer characteristics, such as what kind of gazetteer can benefit the model well, which can bring new insights for us to build a more suitable gazetteer.

\section*{Acknowledgement}
%We thank the anonymous reviewers for their helpful comments on this paper.
This work was supported by the NSFC projects (No.~62072399, No.~U19B2042, No.~61402403), Chinese Knowledge Center for Engineering Sciences and Technology, MoE Engineering Research Center of Digital Library, and the Fundamental Research Funds for the Central Universities (No.~226-2022-00070).

\end{document}